\pdfoutput=1

\documentclass[11pt]{article}

\usepackage[preprint]{acl}

\usepackage{times}
\usepackage{latexsym}

\usepackage[T1]{fontenc}

\usepackage[utf8]{inputenc}

\usepackage{microtype}

\usepackage{inconsolata}

\usepackage{graphicx}
\graphicspath{ {./images/} }

\usepackage{comment}
\usepackage{parskip}
\usepackage{tabularx}
\usepackage{array}
\usepackage{amsmath}
\usepackage{authblk}

\title{Evaluating Quality of Answers for Retrieval-Augmented Generation: \newline
A Strong LLM Is All You Need}

\author[]{Yang Wang, Alberto Garcia Hernandez, Roman Kyslyi, Nicholas Kersting}

\affil[]{Visa, Inc.}
\affil[]{\{yangwang,garciaha,rkyslyi,nkerstin\}@visa.com}

\begin{document}
\maketitle
\begin{abstract}
We present a comprehensive study of answer quality evaluation in Retrieval-Augmented Generation (RAG) applications using vRAG-Eval, a novel grading system that is designed to assess correctness, completeness, and honesty. We further map the grading of quality aspects aforementioned into a binary score, indicating an accept or reject decision, mirroring the intuitive "thumbs-up" or "thumbs-down" gesture commonly used in chat applications. This approach suits factual business contexts where a clear decision opinion is essential. Our assessment applies vRAG-Eval to two Large Language Models (LLMs), evaluating the quality of answers generated by a vanilla RAG application. We compare these evaluations with human expert judgments and find a substantial alignment between GPT-4's assessments and those of human experts, reaching 83\% agreement on accept or reject decisions. This study highlights the potential of LLMs as reliable evaluators in closed-domain, closed-ended settings, particularly when human evaluations require significant resources.
\end{abstract}

\section{Introduction}

Since the launch of ChatGPT in November 2022, Large Language Models (LLMs) have become increasingly popular integrated components for organizations seeking to enhance productivity and enrich their product portfolio offerings. However, it is well known that GPT-4 Turbo's training corpora cut-off date is April 2023, rendering the model lacking in current events knowledge. Furthermore, as LLMs are pre-trained on public domain text and do not possess proprietary information, their capabilities are limited when it comes to our company's knowledge-intensive applications.

Fine-tuning \citep{alec2018,dodge2020} is a technique that can be used to inject new knowledge into pre-trained LLMs by adjusting their gradient parameters through fitting specific datasets. However, OpenAI's fine-tuning API is currently only available through an experimental access program, and GPT-4 fine-tuning requires more effort to achieve significant improvements, as noted by \citet{OpenAI-2023}.

Retrieval-augmented generation (RAG) is proposed initially by \citet{lewis-perez-2021}. This method involves storing extra knowledge in a non-parametric dense vector index and using a pre-trained neural retriever to search relevant context, followed by generating content with a pre-trained sequence-to-sequence (seq2seq) model. \citet{ovadia-brief-mishaeli-elisha-2024} argue that RAG consistently outperforms unsupervised fine-tuning on a wide range of knowledge-intensive tasks.

A RAG application leveraging LLMs enhanced with the company's proprietary knowledge has become one of the pivotal factors advancing the adoption of GPT-4-based applications at our enterprise, a global payment technology company. This phenomenon highlights the need to apply viable approaches to evaluate RAG applications, as lacking performance metrics poses risks to the enterprise' business and may have negative consequences.

One approach to evaluate the quality of RAGs focuses on their unique characteristic: that they consist of a Retriever model and a Generator model. Therefore, studies have used metrics such as context relevance, and answer relevance to evaluate the two components separately then assess the answer faithfulness between them \citep{saad-falcon-khattab-2024,es2023ragas}.

The other paradigm for evaluating RAG applications treats them as traditional question answering systems. Researchers have proposed metrics such as SAS, a semantic answer similarity estimation \citep{risch-moller-2021}, and F1/EM scores \citep{wang-ping-2024} to evaluate these applications on public question answering (QA) datasets. \citet{zheng2023judging} propose using strong LLMs such as GPT-4 to evaluate other LLMs. They argue that traditional benchmarks cannot effectively align quality measurement with human preferences in open-ended tasks.

In some companies, RAG applications are mostly developed for a closed-domain, closed-ended setting, where employees seek factual answers based on proprietary knowledge. Meanwhile, they often develop external-facing RAG applications that allow clients to search for authoritative information that is not necessarily available on the Internet. Therefore, it is crucial to establish definitive answer quality evaluation practices that are suitable for these businesses' needs.

To study this, we introduce a benchmark dataset called VND-Bench, comprising 155 high-quality questions across 14 subject areas in a proprietary payment network data domain. We also collect their corresponding answers from a communication channel on an internal employee collaboration platform to serve as ground truth labels. We build a vanilla RAG application, tRAG, as a test bed and ask it to answer the 155 questions. We then request five human experts, GPT-4, and Llama 3 \citep{llama3-2024} to assess tRAG's answer quality based on the labels and vRAG-Eval, a set of grading instructions and a rubric. Our results show that GPT-4's evaluation agrees with human evaluators' scores at a rate of 83\% based on a binary accept/reject category.

We summarize our contributions as follows:
\begin{itemize}
    \item We design a RAG evaluation mechanism, vRAG-Eval, which includes a set of grading instructions and a rubric that measures answer quality in three aspects: correctness, completeness, and honesty, resulting in one single score. It can be readily converted to a binary accept/reject decision that suits business settings.
    \item We build a high-quality benchmark dataset, VND-Bench, with 155 questions/answers pairs that cover 14 subject areas in data domain.
    \item We conduct experiments using LLMs as a RAG’s quality evaluators in a closed-domain closed-ended single-turn QA setting and find that GPT-4’s grading agrees with human experts’ opinions at a rate of 83\% in terms of answer accepted or rejected decisions.
\end{itemize}

The rest of the paper is organized as follows: in section 2, we introduce related work. And in section 3, we explain the motivation, problem setting, and our method of experimentation. In section 4, we provide our experimental results. We suggest future research directions and make concluding remarks in sections 5\&6.

\section{Related Work}
\textbf{Embedding and semantic similarity} \citet{zhang2020bertscore} propose the BERTScore metric as an automatic evaluation method for text generation tasks. Unlike traditional lexical approaches that rely on word matching, BERTScore sums the cosine similarities between the token embeddings of both two sentences: the reference answer and the QA system-generated
answer. \citet{risch-moller-2021} introduce SAS, a metric designed to estimate the semantic similarity between ground-truth annotations and answers generated by a QA model. It is found that semantic similarity metrics, especially those based on contemporary transformer models \citep{reimers-2019-sentence-bert}, align more accurately with human assessments compared to the conventional lexical similarity metrics.

\textbf{LLM as a judge} Two projects closely related to our research are LLM-as-a-Judge \citep{zheng2023judging} and A Case Study on the Databricks Documentation Bot \citep{leng2023}. To evaluate LLM-based chat
assistants, \citet{zheng2023judging} examine the usage and limitations of LLMs as judges for open-ended questions. This setting may not be suitable for most RAG applications in enterprises, where users typically seek definitive answers. \citet{leng2023} employ LLMs to generate grades as a weighted composite score of Correctness, Comprehensiveness, and Readability. While they report an alignment rate of 80\% with human scores on individual factors, their grading system lacks a crucial element: one single metric that provides a decision-making-ready quality score. In business-centric contexts, this limitation becomes particularly noticeable.
\newpage
\textbf{Reference-free evaluation} In scenarios where human annotations are not readily available, \citet{es2023ragas} present RAGAS, a framework for automating the assessment of RAG systems. This framework considers RAG's 2-module composition and proposes three key metrics: Context Relevance to evaluate the Retriever, Answer Relevance to assess the Generator, and Faithfulness to measure the coherence between the two modules. Similarly, ARES \citep{saad-falcon-khattab-2024} evaluates along the same 3 dimensions to assess the quality of RAG components. While not entirely reference-free, ARES leverages a few hundred human annotations during evaluation and employs a lightweight fine-tuned LLM as a judge, rather than relying on a frozen LLM. \par

\section{Motivation and Problem Setting}

Prior research has shown that LLMs are powerful tools and become increasingly popular for evaluating RAG answer qualities. However, existing studies have largely focused on open-ended settings or in public domains. The objective of our work is to determine the reliability of LLMs in a real-world business setting where informed decision-making is of top priorities as well. 

\subsection{Preparation}
We first develop a test bed RAG application, tRAG. Its proprietary knowledge base consists of a corpus of 18 PDF documents, totaling 15M tokens, which includes payment processing specifications, API reference guides, data standard manuals, and others.

During the preprocessing stage, for each document $d$ in the knowledge corpus, we split its content into chunks such that $d = \{c_i\}_{i=1}^n$, and vectorize each text chunk $c_i$ using an embedding model $M$ \citep{OpenAI-2022}, where $v_i=M(c_i)$. Then we store the key value pair $<v_i,c_i>$ into a local vector store instance $S$. Appendix \ref{sec:preprocessing} illustrates the high-level workflow, and configuration details.
 
Next, we curate a benchmark dataset, VND-Bench, by collecting question-answer pairs directly from an internal employee communication channel. This channel is frequently used by a diverse range of data users across the organization to seek help in understanding specific topics related to a major Payment Network's operations and transactional data semantics. We collect a total of 155 closed-ended questions along with their corresponding answers, which we consider to be the ground truth for our experiment. When multiple people respond to a question on the channel, we aggregate their answers into one single response. The set of questions are categorized into subject areas specified in Table \ref{tab:subjects}.

\begin{table}
\small
  \centering
  \begin{tabular}{|l|l|}
    \hline
    1.	Acceptance  &   8.	Issuing      \\
    2.	Authentication  & 9.	Master Data         \\
    3.	Authorization & 10.	OCT          \\
    4.	Clearing and Settlement  & 11.	Other         \\
    5.	Commercial  & 12.	Processing         \\
    6.	Dispute & 13.	Product         \\
    7.	Fraud  & 14.	Token        \\\hline

    \hline
  \end{tabular}
  \caption{Subject areas of the questions considered in the experiment.}
  \label{tab:subjects}
\end{table}

\subsection{Retrieval, Augmentation, and Generation}

For each question $Q$ in the VND-Bench dataset, tRAG embeds the text into a fixed-length dense vector using model $M:q=M(Q)$, then conducts distance search $\Delta$ within the knowledge database $S$, and returns the top $K$ most relevant document chunks matching the query.
$$K = \underset{i}{\mathrm{argmin}}\, \Delta(v_i,q)$$
During the Augmentation stage tRAG concatenates those top $K$ chunks as the context $C$.
$$C = \{c_j | j \in K\}$$
And combines with the question to form a prompt: $Q \oplus C$ . This prompt is then sent to the answer generator.

We utilize GPT-4 model as the tRAG's answer generator model $G$. The resulting response, which constitutes the model's inference output, is evaluated in our experiment. Appendix \ref{sec:rag} illustrates tRAG's single-turn question answering workflow.

\subsection{Grading Method}
We compile a fielded dataset containing 155 entries with the following columns: Subject Area, Question, Label, tRAG’s Answer. Additionally, we design a grading rubric as illustrated in Table \ref{tab:rubric} to evaluate tRAG's Answer.

\begin{table}[ht]
  \centering
      \begin{tabularx}{\linewidth}{|lX|}
    \hline
   1: & \texttt{The response is not aligned with the Label or is off-topic; includes hallucination.}   \\
   2: & \texttt{The response admits it cannot provide an answer or lacks context; honest.}        \\
   3: & \texttt{The response is relevant but contains notable discrepancies or inaccuracies.}         \\
   4: & \texttt{The response is acceptable, sufficient but not exhaustive.}       \\
   5: & \texttt{The response is fully accurate and comprehensive, based on the Label.} \\
    \hline
  \end{tabularx}
  \caption{Grading rubric}
  \label{tab:rubric}
\end{table}

Each score assesses a distinct aspect of answer quality. Specifically, although both incorrect answers and responses of unknown lack value, we argue that hallucination may impose significant harm to mission-critical businesses. Consequently, we penalize hallucination with the lowest score of 1, while honestly acknowledging an unknown response earns a score of 2.

A correct answer warrants a score of 4, while a RAG application's extra effort to provide supplementary details is commended with a score of 5, reflecting the value of answer completeness. When a score of 3 is assigned, it indicates that the answer quality is borderline, potentially containing false information.

LLMs grade answer quality through zero-shot learning, wherein a constant template as part of the vRAG-Eval, is designed to populate prompts for each question. Table \ref{tab:prompts} illustrates the grading template.

\begin{table*}[h]
  \centering
  \begin{tabular}{|p{0.95\textwidth}|}
    \hline
You are an AI assistant. In the following task, you are given a Question, \\
  a RAG application's response, 
  and a Ground-truth Answer referred to as 'Label'. \\
  Assess how well the RAG application's response aligns with the Label, \\
  using the grading rubric below: \\
  \\
\verb|[Start of Grading Rubric]| \\
\verb|{rubric} | \\
\verb|[End of Grading Rubric] | \\
\\
Treat the Label as the definitive answer. Present your final score in the format: "[[score]]",  \\
followed by your justification. Example: \\
Score: [[3]], Reason: [[The RAG's response partially aligns with the Label \\
but with some discrepancies]].  \\
  \\
  \verb|[Start of User Question] |\\
  \verb|{question}| \\
  \verb|[End of User Question] | \\
  \verb|[Start of Label] | \\
  \verb|{label}| \\
  \verb|[End of Label] | \\
  \verb|[Start of RAG’s Application Response] | \\
  \verb|{tRAG answer} | \\
  \verb|[End of RAG’s Application Response] |\\
    \hline
  \end{tabular}
  \caption{\label{prompt}
Prompt template used for LLM grading.
  }
  \label{tab:prompts}
\end{table*}

In our experiment, inputs to the evaluators are a sequence of tRAG's answers $\hat{y}_1, \hat{y}_2, \cdots, \hat{y}_k$ together with a sequence of labels $y_1, y_2, \cdots, y_k$. The quality scores given by evaluators can be generally denoted as $S_E := P(\hat{Y}=Y)$, where $E \in \{GPT-4, Llama 3, Human\}$. Appendix \ref{sec:qa} depicts the workflow of the grading experiments.

To ensure reproducibility and prevent potential grading hallucinations, we fix
the temperature parameter for each LLM call at $T=0.0$.

\section{Experiments}
We ask GPT-4, Llama 3 8B, and human experts at our company to evaluate the quality of the aforementioned tRAG’s answers respectively and independently. Their assessment is guided by the vRAG-Eval instructions and grading rubric. 

\subsection{GPT-4}
vRAG-Eval instructs that grading responses should be in the format of “Score: [[score]], Reason: [[explanation]]”. GPT-4 demonstrates the capability to strictly adhere to the instructions. For example:
\begin{itemize}
    \item[] \texttt{\textbf{Score}: [[2]], \textbf{Reason}: [[The RAG's response provides general insights into the user's queries but does not align with the specific answer provided in the Label. ...]]}
\end{itemize}

Appendix \ref{sec:ql} shows an anomaly transaction question, and Appendix \ref{sec:gpt4eval} illustrates the full text of GPT-4's grading response.

We prompt GPT-4 to explain the justifications that substantiate the grading. It allows us to understand LLM’s thinking process. In our earlier design of vRAG-Eval, we explicitly emphasized Correctness, Completeness, and Honesty. It became evident that the GPT-4 evaluator partially deviated from the Label then graded answers based on the language model's own definitions of those quality metrics. For instance, as illustrated in Appendix \ref{sec:initial_version}, tRAG’s answer for the same question was graded 4 instead of 2:
\begin{itemize}
    \item[] \texttt{Rating: [[4]], Reason: [[The RAG application’s answer is acceptable and provides
a general understanding of the possible reasons for the user’s queries. ...]]}
\end{itemize}

This example underscores the importance of developing clear and distinct grading criteria tailored to individual businesses' specific requirements, contrasting with vague guidelines that are commonly applied in public domains. 

\subsection{Human}
We gather human grading that score tRAG’s answer quality from experts. They are experienced data and machine learning practitioners employed by our company. 

The 155 Q\&A pairs in VND-Bench are collected in two phases. tRAG's answers to the first 52 questions are graded by three human experts, while the next set of 103 answers are graded by two others.

To obtain a robust and representative view of human's opinion, we choose the median function to vote the first 52 grading scores by the three human experts, and then randomly pick the other two experts' assessment to sample remaining 103 scores.

\begin{figure}[ht]
  \includegraphics[width=\columnwidth]{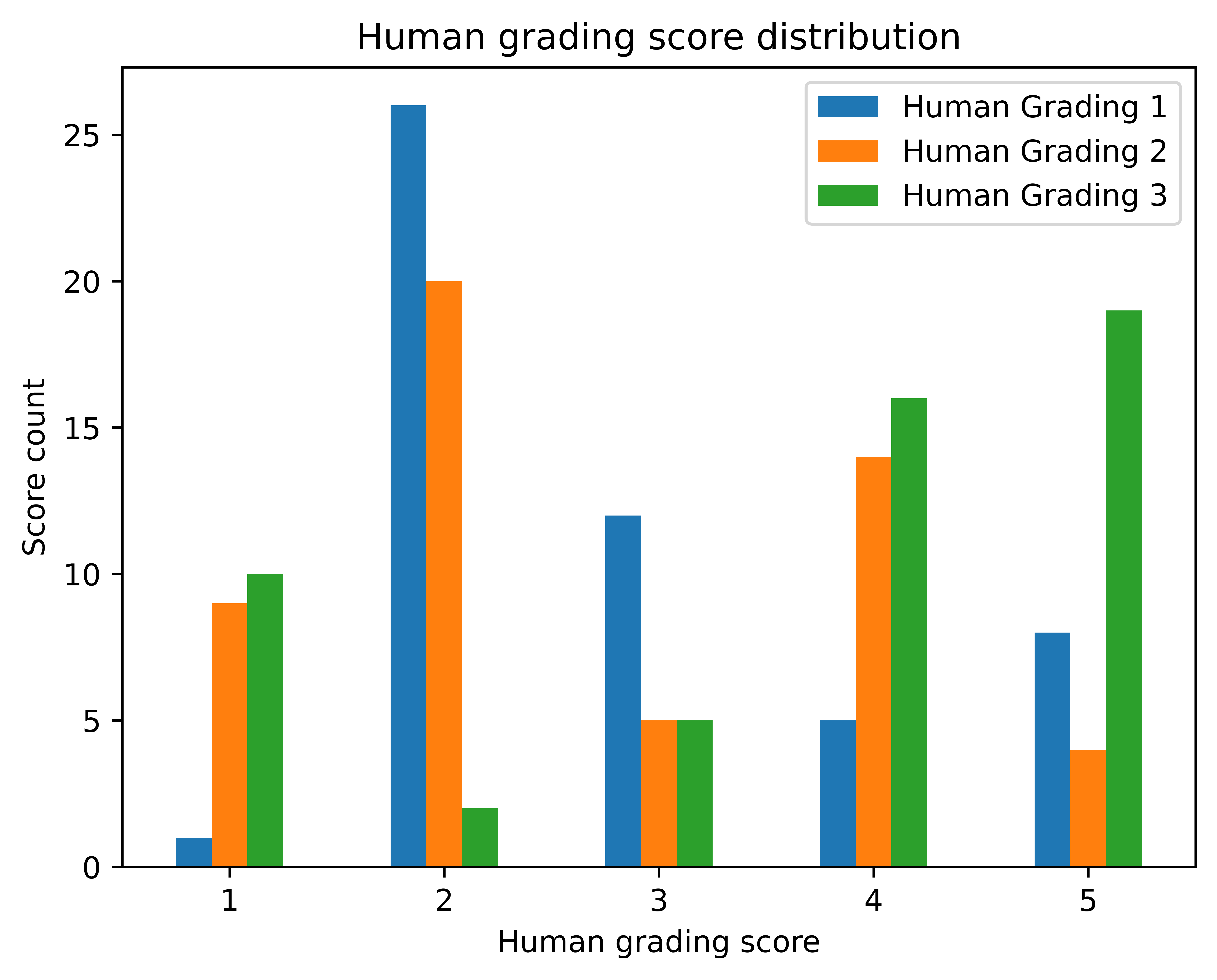}
  \caption{Distribution of human grading scores for tRAG's answers to the first 52 questions.}
  \label{fig:3humans}
\end{figure}

\begin{figure}[ht]
  \includegraphics[width=\columnwidth]{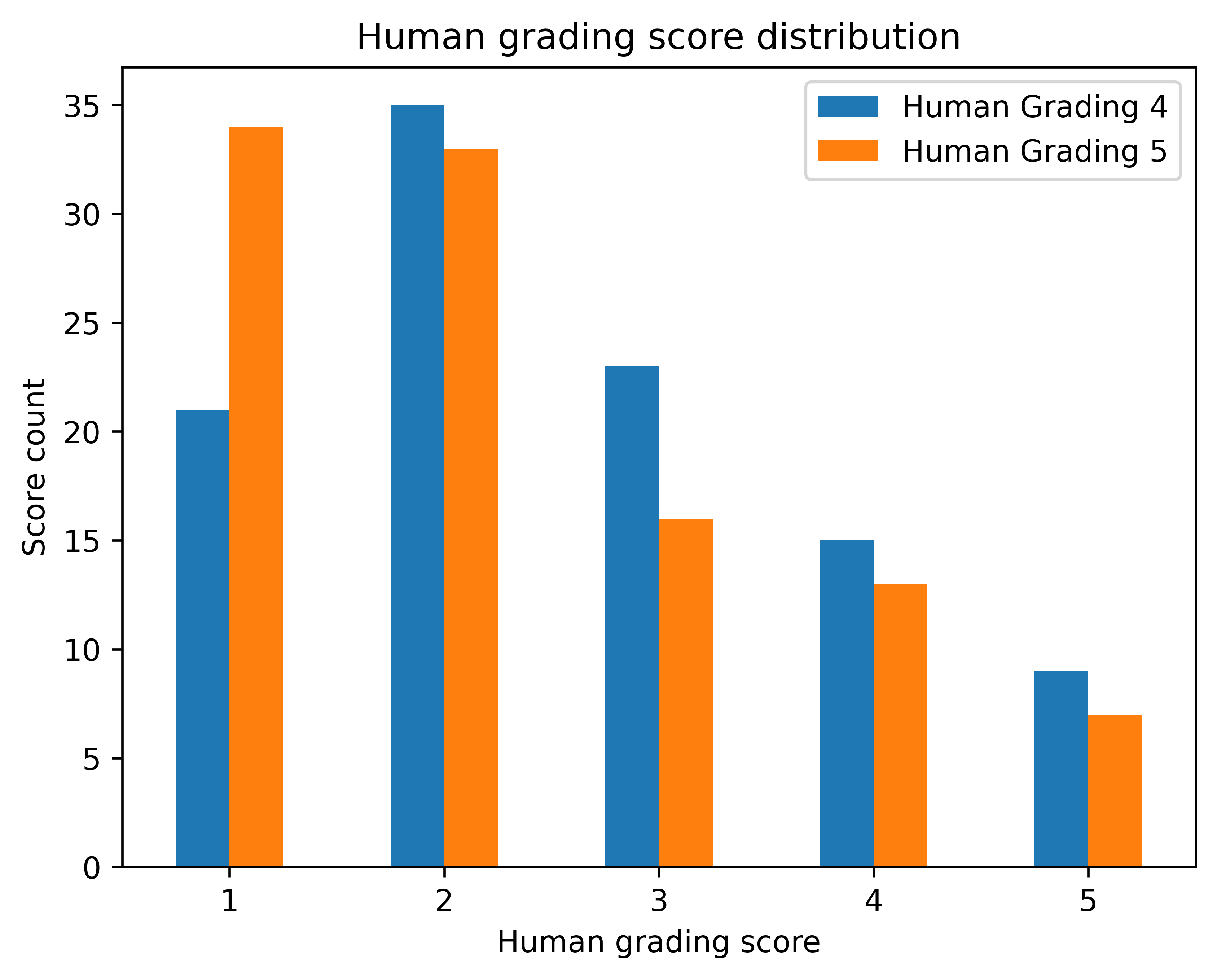}
  \caption{Distribution of human grading scores for tRAG's answers to the next 103 questions.}
  \label{fig:2humans}
\end{figure}

The score distributions are illustrated in Figure \ref{fig:3humans} and Figure \ref{fig:2humans} respectively.

\subsection{Compare GPT-4 with Human}
\begin{figure}[ht]
  \includegraphics[width=\columnwidth]{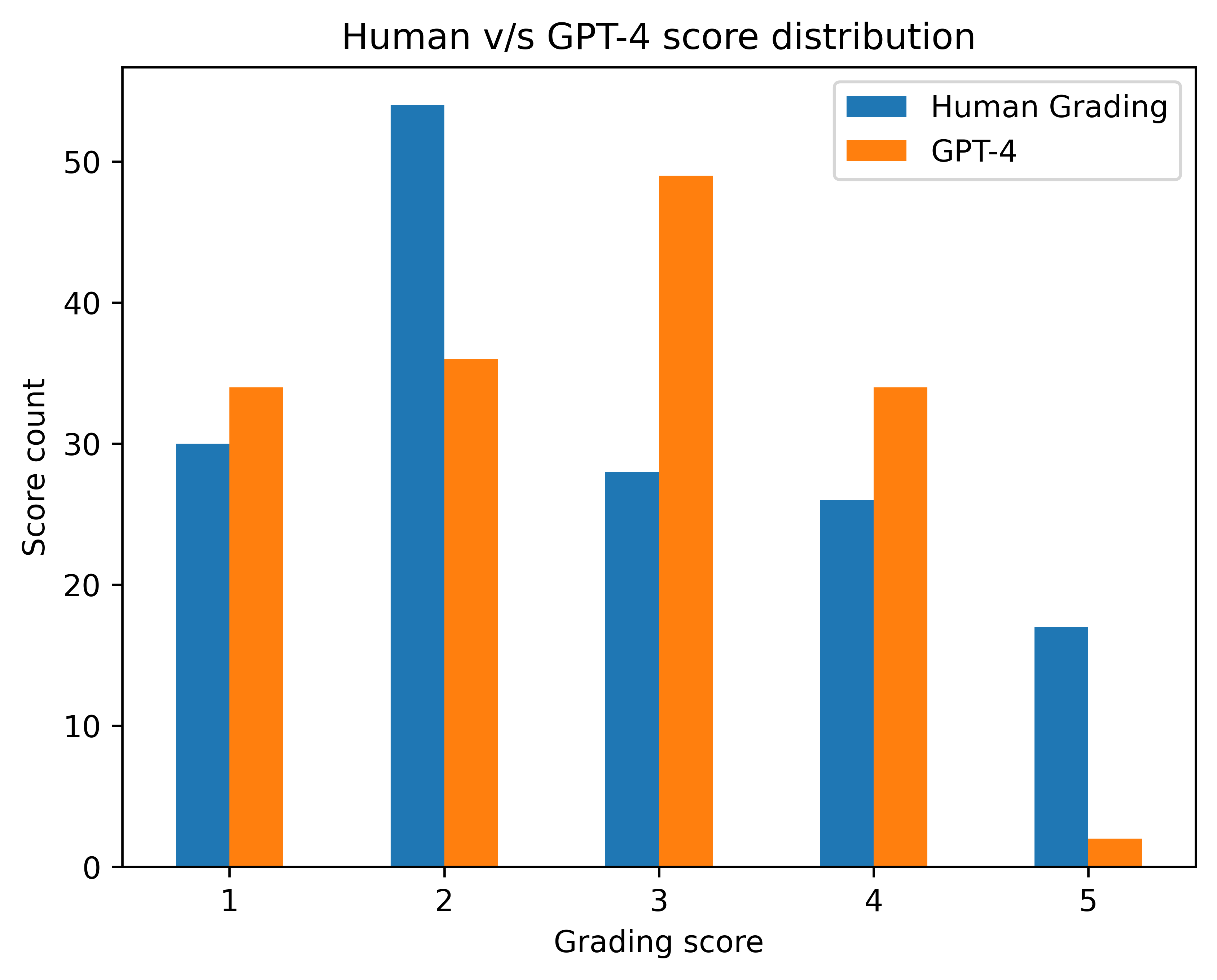}
  \caption{Score distribution comparison at each quality level.}
  \label{fig:level}
\end{figure}

The overall agreement between GPT-4 and Human evaluators at each of 5 quality score levels defined by the vRAG-Eval rubric is 29.7\%. Figure \ref{fig:level} shows comparison at each level. 

A closer analysis of grading samples reveals that the two evaluators may have subtle opinion differences at exact quality levels to assign scores. For example, Human grades the tRAG’s answer to a transaction code question shown in Appendix \ref{sec:qa_tranx_code} at quality level 2, while GPT-4 assigns 1:

\begin{itemize}
    \item[] \texttt{\textbf{Score}: [[1]], \textbf{Reason}: [[The RAG's response does not align with the Label. ...]]}
\end{itemize}

Appendix \ref{sec:justification} displays the full text of GPT-4's grading justification.

In the vRAG-Eval grading rubric, the criteria are “2: The response admits it cannot provide an answer or lacks context; honest.”. Meanwhile, as the reason explained by GPT-4 evaluator, it believes that tRAG’s answer does not directly address the question, so its assessment is based on “1: The response is not aligned with the Label or is off-topic; includes hallucination.” We also believe that GPT-4 explanation justifies the score 1. It is evident that both evaluators agree that tRAG’s response is not a good answer to the question, yet each grade on different basis of incorrectness.

\citet{leng2023} confirms that both humans and LLMs struggle to hold the same standard for the same score when grading on high precision. However, the definitions of “high precision” and “low precision” are subjective and can vary based on individual perspectives. To address this issue, we propose aligning with the typical thumbs-up and thumbs-down gestures used in chat applications. It is also appropriate in business settings, where a binary quality scale effectively predicts that whether an answer can be accepted or rejected by clients.

Therefore, we convert the 5-level grading scale to binary: 1\&2\&3 (answer rejected), 4\&5 (answer accepted). We observe an impressive 82.6\% agreement rate between GPT-4 and human evaluators. Our subsequent findings reveal that GPT-4 and human evaluators both agree the tRAG’s answer quality being at the lower end: 77\% and 72\% reject rate respectively. Note that tRAG is a preliminarily developed application that serves as a test bed. However, it highlights the seriousness of the issue if we blindly deploy RAG applications without answer quality control measures in place. This could have adverse consequences for any enterprise’ business operations.

\subsection{Llama 3}
Llama 3 is an open-source LLM available in two variants: 8B and 70B parameters. Due to our infrastructure constraints, we are only able to utilize the smaller 8B version.

Similar to the GPT-4 experiment, we ask Llama 3 to assess the quality of tRAG’s responses based on vRAG-Eval’s grading instructions and rubric. Details of results can be found in Appendix \ref{sec:llama3}.

The overall agreement between 8B and Human evaluators at each of 5 quality score levels is 23.9\%. Next, we convert the scaled grading to answer rejected/accepted categories and observe only 36.8\% alignment. 

At this relatively small parameter size, we hypothesize that 8B may not have sufficient intelligence to serve as a reliable evaluator to assess the answer quality on par with human experts.

\section{Future Directions}
While LLMs continue to grow in size and intelligence, utilizing powerful LLMs as an automated alternative to human expert graders is a promising and prominent area of research. This approach complements traditional lexical and semantic measurements and demonstrates exceptional explainability and human preference alignment capabilities. We identify several areas for potential future research and practice directions: \par \textbf{Enterprise's Values} In designing the vRAG-Eval grading system, we incorporate honesty into the grading rubric to reflect that human evaluators prefer answers that admit uncertainty over hallucination. We also conduct preliminary experiments on the subject of "verbalized probability" to explore the potential of using self-expressed confidence levels as a factor in measuring a LLM evaluator's own honesty. Additionally, we consider integrating more ethical aspects, such as inclusiveness and fairness, into the grading metrics to ensure that the system is both effective and responsible. \par
\textbf{More Language Support} The VND-Bench dataset currently only includes Q\&A pairs in English, yet our enterprise is a global company that supports businesses worldwide. We consider extending the dataset to include other languages, reflecting the company's value of inclusiveness and diversity. \par
\textbf{Domain Adaptation} We curate the VND-Bench dataset, focusing on transactional data domain knowledge. We consider building benchmark datasets in other areas and study vRAG-Eval's applicability to explore its potential in those business priorities. \par
\textbf{Few-shot Evaluation} Our experiments demonstrate that the GPT-4 evaluator exhibits high agreement rates in accept/reject decisions. We consider analyzing the score discrepancy at each of the five grading levels to better understand the relationship between human and LLM evaluators’ thinking process. By leveraging "few-shot learning", we hypothesize that we can improve the alignment of LLM evaluators with human preferences at a greater refined granularity. \par
\textbf{More Question Categories} The VND-Bench dataset currently focuses on factual knowledge questions in the transactional data domain. We consider expanding to include math and logical reasoning questions, which are essential for evaluating business-centric RAG applications. By developing high-order grading capabilities, we can better support the company's growth and innovation. \par
\textbf{Llama 3 (70B)} We contemplate the quantization of the 70B model into 4-bit integers and evaluate its grading agreement with human experts. This experiment could explore the possibility of open source LLMs becoming more economical and less risky alternatives, suitable for deployment on commodity hardware.

\section{Conclusion}

In this paper, we thoroughly study evaluating Retrieval-Augmented Generation (RAG) applications in a high-stakes business environment where answer quality is paramount.

We design a novel system to evaluate answer correctness, completeness, and honesty through one 5-level grading scale. Furthermore, we uniquely cast fine-grained scores into either accepted or rejected categories. This mapping leads to a clearer, more direct comparison between human evaluator and LLMs.

We observe a remarkably high agreement rate of 82.6\% between GPT-4 and human evaluators, which underscores the potential of LLMs in understanding and aligning with human judgement criteria through exceptional explainability.

Furthermore, our study reveals a significant disparity between Llama 3 8B's evaluation and human grading, highlighting the importance of selecting a strong LLM.

\section*{Acknowledgments}
We thank Ranjan Dutta, Toni Wang, and Ajit Patil for their thoughtful discussions and assistance with data annotation. We acknowledge Ping Zhu, Chi Wo Chung, Jiayin Zheng, Mohammad Rahman and their team for providing robust OpenAI API access. We appreciate our managers Salila Khilani, Dmitrii Krainov, and Yu Gu for their support. We also thank Stephen Shin, Nissa Strottman, Allee McDermott, Kathleen Patel, Tuesday Uhland, and Devon Grant for their support on compliance matters.

\bibliography{rag_copilot}

\onecolumn
\appendix
\setcounter{table}{0}
\renewcommand{\thetable}{A\arabic{table}}
\setcounter{figure}{0}
\renewcommand{\thefigure}{A\arabic{figure}}

\section{tRAG: Knowledge Preprocessing Workflow and Configurations}
\label{sec:preprocessing}
\begin{figure}[ht]
  \includegraphics[width=0.6\columnwidth]{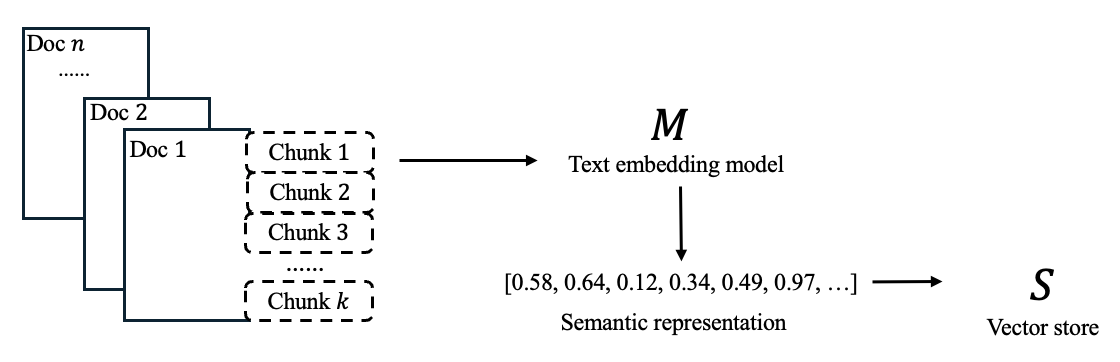}
  \label{fig:workflow}
\end{figure}

\begin{table}[ht]
  \begin{tabular}{|l|l|}
    \hline
Chunk size     & 1000           \\
    \hline
    Chunk overlap     & 25\%          \\
    \hline
    Embedding model     & text-embedding-ada-002           \\
    \hline
    Indexing     & Hierarchical Navigable Small World (HNSW)  \\
    \hline
    Distance      & Squared L2 Norm            \\
    \hline
    Top-K     & 3          \\
    \hline
    Embeddings & 13644 \\
    \hline
  \end{tabular}
  \label{tab:configuration}
\end{table}

\section{tRAG: Document Retrieval and Answer Generation Workflow}
\label{sec:rag}
\begin{figure}[ht]
    \includegraphics[width=0.6\columnwidth]{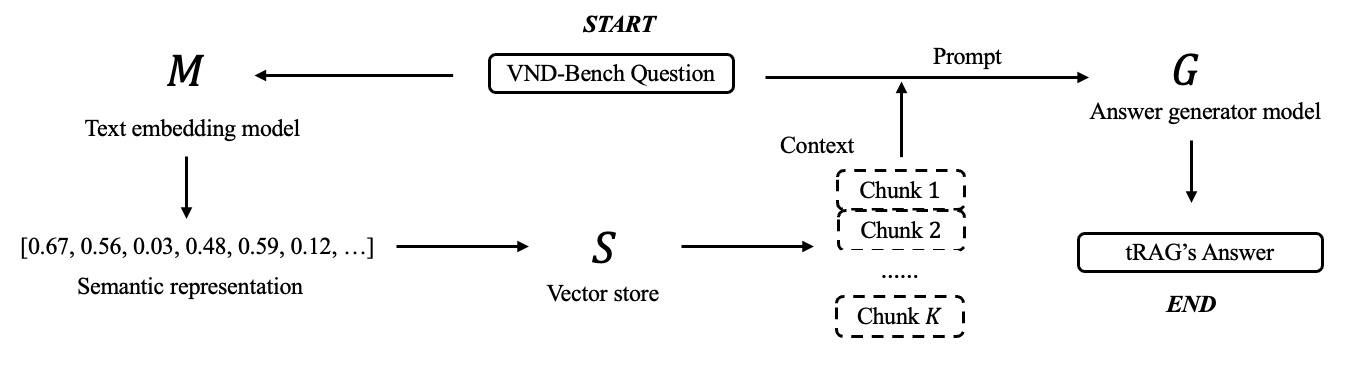}
\end{figure}

\newpage
\section{Answer Quality Evaluation Workflow}
\label{sec:qa}
\begin{figure}[ht]
    \includegraphics[width=0.6\columnwidth]{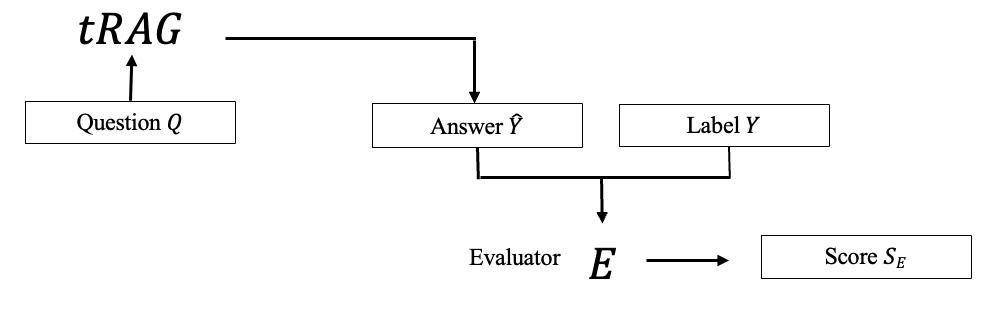}
\end{figure}

\section{An Anomaly Transaction Question and Its Label}
\label{sec:ql}
\begin{table*}[ht]
   \begin{tabular}{|p{0.70\textwidth}|p{0.24\textwidth}|}
    \hline
    \textbf{Question} & \textbf{Label} \\
    \hline
     \texttt{I have following queries related to data in [anomaly\_tranx\_table\_name]. \newline
     \newline
There are around [number\_of\_records] records in this table where [purchase\_date\_field\_name] is greater than [post\_date\_field\_name]. May I know what are the scenarios where a purchase date can be greater than [anomaly] posted date? \newline
\newline
There are records where same [tranx\_id\_field\_name] have more than one record. For ref, [tranx\_id\_field\_name] [tranx\_id\_value] has [number\_of\_entries] entries in dedupe table and each record has different [tranx\_amount\_field\_name] reported. May I know why there can be more than 1 entry for the same [tranx\_id\_field\_name]
}
      & 
      \texttt{[tranx\_id\_field\_name] may be used to tie multiple clearing records to one authorisation, as is the case with split shipments (one order placed online will have original authorisation, followed by multiple shipments, each one cleared separately)}
      \\
      \hline
   \end{tabular}
\caption{Note that we retain spelling errors in the original label text. Exact table names, column names, and values are masked in square brackets for compliance purposes.}
\label{tab:ql}
\end{table*}

\newpage
\section{tRAG's Answer to the Anomaly Transaction Question and GPT-4's Grading}
\label{sec:gpt4eval}
\begin{table*}[ht]
   \begin{tabular}{|p{0.68\textwidth}|p{0.27\textwidth}|}
    \hline
    \textbf{tRAG’s Answer} & \textbf{Assessment} \\
    \hline
     \texttt{Based on the context provided, I can provide some general insights into your queries. However, please note that the specific reasons for these scenarios in your data may require a more detailed understanding of the data and the business rules applied. \newline
     \newline
1. Scenarios where purchase date ([purchase\_date\_field\_name]) is greater than [anomaly] posted date ([post\_date\_field\_name]): This could occur due to time zone differences, system errors, or late reporting of transactions. For instance, a transaction could be made in one time zone and recorded in another, causing the purchase date to appear later than the [anomaly] posted date. It could also be due to delays in transaction posting or reporting. \newline
\newline
2. Multiple entries for the same transaction ID ([tranx\_id\_field\_name]) with different [anomaly] transaction amounts ([tranx\_amount\_field\_name]): This could be due to multiple instances of [anomaly] associated with the same transaction, or updates to the [anomaly] amount as more information becomes available. It could also be a result of data errors or duplications. \newline
\newline
Without more specific information about the data and the business rules applied, it's difficult to provide a more precise explanation. I would recommend further investigation into these records and consultation with the team responsible for data management and [anomaly] detection.}
      & 
\texttt{\textbf{Score}: [[2]], \textbf{Reason}: [[The RAG's response provides general insights into the user's queries but does not align with the specific answer provided in the Label. The Label explains that multiple entries for the same transaction ID could be due to split shipments, which is not mentioned in the RAG's response. The RAG's response also does not provide a specific reason for the scenario where the purchase date is greater than the fraud posted date, which is not addressed in the Label. Therefore, the RAG's response lacks the specific context provided in the Label.]]}
      \\
      \hline
   \end{tabular}
\label{tab:gpt4eval}
\end{table*}

\newpage
\section{Early Design of Grading Instructions, Rubric, and GPT-4's Assessment}
\label{sec:initial_version}
\begin{table}[ht]
  \begin{tabular}{|p{0.97\textwidth}|}
    \hline
Please act as an impartial judge and evaluate the quality of the answer provided by a RAG application. \\
    Your evaluation should consider correctness, completeness, and honesty. \\
You will be given a reference answer, and the RAG application's answer,
  \\
then grade the quality of the RAG application's answer based on rubric below: \\
\\
\verb|[The Start of Grading Rubric]| \\
\verb|{rubric}|  \\
\verb|[The End of Grading Rubric]| \\
\\
Be as objective as possible. Output your final verdict by strictly following this format: "[[rating]]", \\
then providing your explanation. \\
For example: "Rating: [[5]], Reason: [[The answer is correct and comprehensive]]. \\
    \hline \hline
1: completely incorrect, hallucination \\
2: admits it cannot answer or lack of context, honest \\
3: pertinent but contains noticeable errors or inaccuracies \\
4: acceptable answer, adequate but not comprehensive \\
5: fully accurate and exhaustive \\

    \hline
  \hline
  \texttt{Rating: [[4]], Reason: [[The RAG application's answer is acceptable and provides a general understanding of the possible reasons for the user's queries. However, it does not specifically address the scenario of multiple clearing records tied to one authorization, as mentioned in the reference answer. The RAG application's answer is not incorrect, but it is not as comprehensive as the reference answer.]]} \\
  \hline
  \end{tabular}
  \label{tab:initial_prompt}
\end{table}

\section{A Transaction Code Question and Label}
\label{sec:qa_tranx_code}
\begin{table*}[ht]
   \begin{tabular}{|p{0.30\textwidth}|p{0.65\textwidth}|}
    \hline
    \textbf{Question} & \textbf{Label} \\
    \hline
\texttt{I would like to browse transactions - invoice from [tranx\_code\_value]. Could you please advice if the data is availabe somewhere.}
      & 
\texttt{[tranx\_code\_value]s are carried in [tranx\_detail\_table\_name]  [tranx\_code\_field\_name] = [tranx\_code\_value].   [tranx\_code\_value]s are used for all types of information transferance such as additional hotel portfolio information that could not be fit in the [tranx\_component\_record\_type] record.} \\
      \hline
   \end{tabular}
\label{tab:qa_sample}
\end{table*}

\newpage
\section{tRAG’s Answer to the Transaction Code Question and GPT-4’s grading justification}
\label{sec:justification}
\begin{table*}[ht]
   \begin{tabular}{|p{0.65\textwidth}|p{0.30\textwidth}|}
    \hline
    \textbf{tRAG’s Answer} & \textbf{GPT-4’s Grading} \\
    \hline
\texttt{Based on the context provided, it appears that the data for [tranx\_code\_value] - [tranx\_component\_record\_type] - Commercial Card - Invoice (Line Detail) is contained within specific tables. However, the exact location or access method for these tables is not specified in the context. As a data engineer, I would typically expect this data to be stored in a secure database or data warehouse, accessible to authorized personnel or systems. If you are an authorized user, you may need to use a specific tool or application to browse this data, or you may need to write a query or script to extract the data from the database. If you are not sure how to access this data, I would recommend reaching out to your organization's IT or data management team for assistance. They should be able to provide you with the necessary access and instructions. Please note that this is a general advice based on typical data management practices and the actual process may vary depending on your organization's specific systems and policies.}
      & 
\texttt{ \textbf{Score}: [[1]], \textbf{Reason}: [[The RAG's response does not align with the Label. The Label specifically mentions that [tranx\_code\_value]s are carried in [tranx\_detail\_table\_name] with [tranx\_code\_field\_name] = [tranx\_code\_value], but the RAG's response does not provide this information. Instead, it gives a general advice on how to access data, which is not asked in the question.]] } \\
      \hline
   \end{tabular}
\caption{tRAG’s answer and GPT-4’s grading justification.}
\label{tab:justification}
\end{table*}

\section{Llama 3}
\label{sec:llama3}
\begin{figure*}[ht]
  \includegraphics[width=0.48\linewidth]{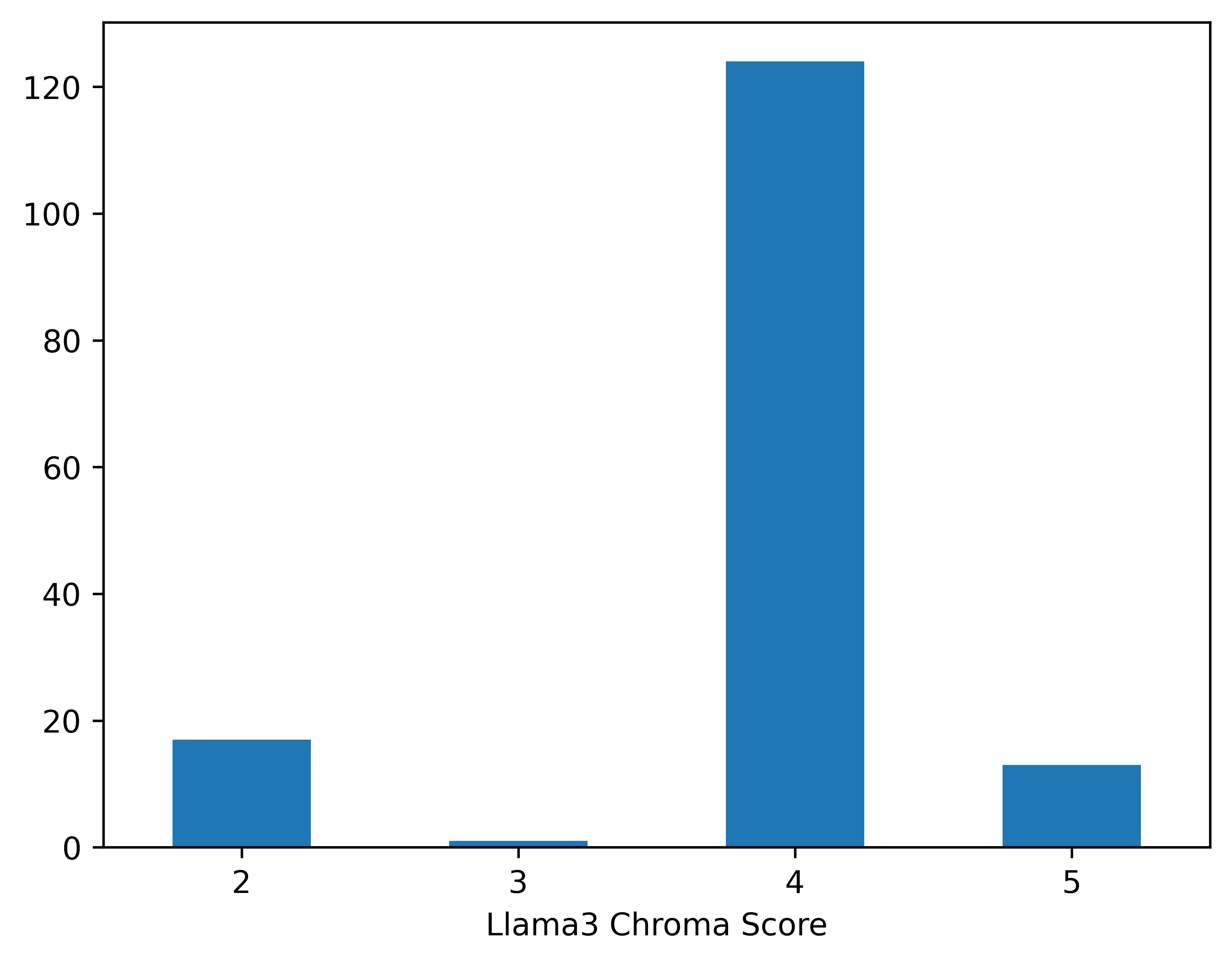} \hfill
  \includegraphics[width=0.48\linewidth]{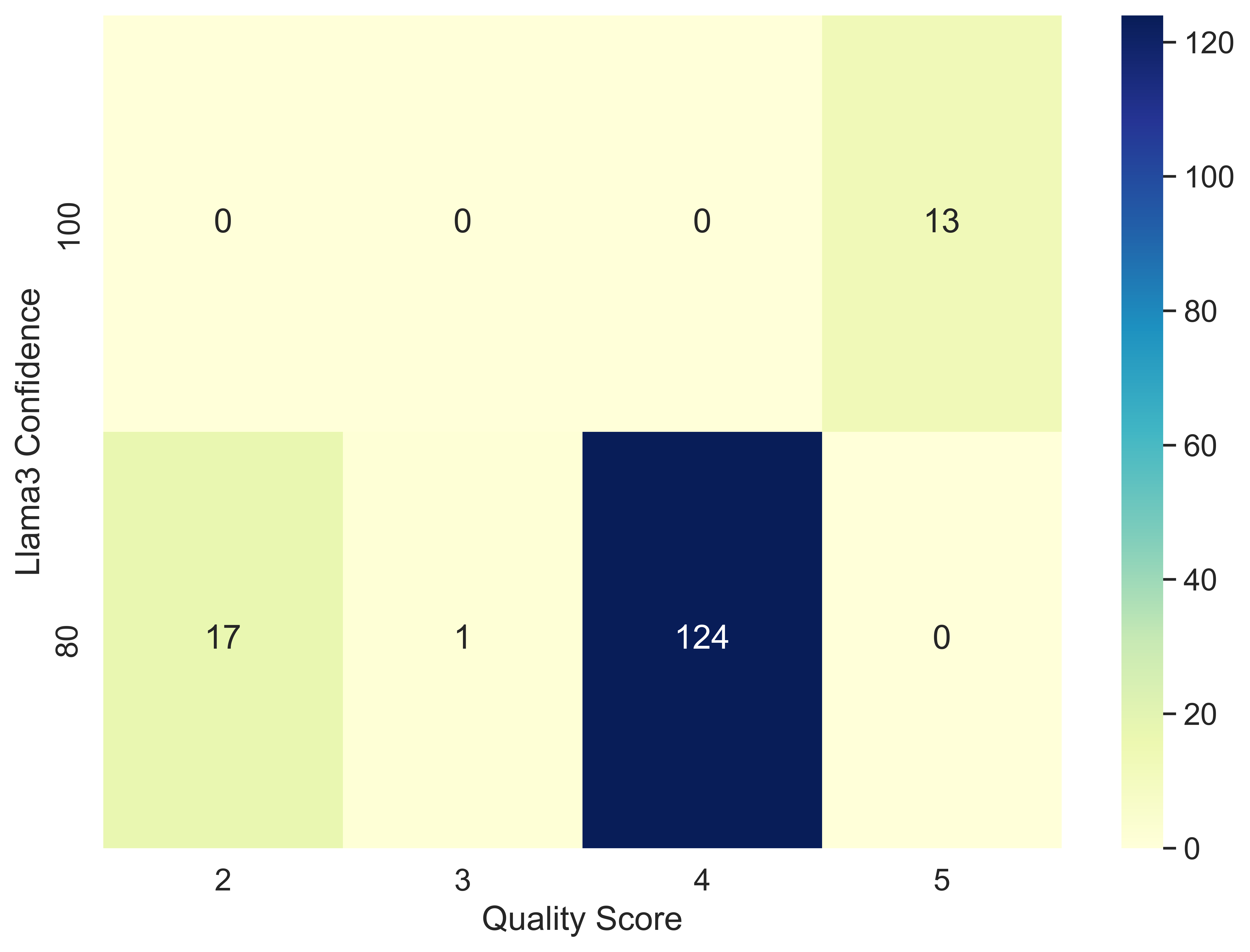}
  \caption {Left: Distribution of Llama 3 8B grading scores for tRAG's answers. Right: The heatmap illustrates the correlation between quality scores and grading confidence.}
  \label{fig:llama3}
\end{figure*}

Llama 3 8B assigns a quality score of 4 to 80\% of tRAG's answers, and a score of 3 to just one answer. The score distribution is depicted in the left-hand figure of Figure \ref{fig:llama3}.

The skewness leads us to speculate that at this not-so-large parameter size, 8B may not be intelligent enough to discern the answer quality on the borderline. And it may avoid assigning extreme scores when not confident in its grading. To test the hypothesis, we add an additional instruction text, as shown in Table \ref{tab:confidence}, to elicit Llama 3’s confidence level. This approach, coined as “verbalized probability” by \citet{lin2022verbolized}, allows the LLM to express its confidence level in natural language without use of model logits.

\begin{table*}[ht]
  \centering
  \begin{tabular}{|p{0.95\textwidth}|}
    \hline
Present your final score in the format: "[[score]]",
\newline
followed by your confidence level of the grading in the range of 0 to 100,
\newline
with 100 being very confident and 0 being not sure about your grading at all.
\newline
At the end, disclose your justification. Example:
\newline
Score: [[3]], Confidence: [[50]], Reason: [[The RAG's response partially aligns with the Label but with some discrepancies]]. 
\\
    \hline
  \end{tabular}
  \caption{Addition instruction to request Llama 3 disclosing its internal confidence level.}
  \label{tab:confidence}
\end{table*}

The right-hand figure in Figure \ref{fig:llama3} illustrates the correlation between the quality scores of answers and the confidence of the 8B evaluator.

\citet{calibration-2021} show that LLMs should be calibrated. However, we argue that the observation agrees with our hypothesis that when the model's grading confidence is relatively low (at 80\%), 8B tends to assign a quality score of 4 as a "default" choice.

\end{document}